# Automatic Segmentation of the Spinal Cord Nerve Rootlets

**AUTHORS:**

Jan Valošek[1,2,3,4], Theo Mathieu[1], Raphaëlle Schlienger[5], Olivia S. Kowalczyk[6,7], Julien Cohen-Adad[1,2,8,9]

**AFFILIATIONS:**

1. NeuroPoly Lab, Institute of Biomedical Engineering, Polytechnique Montreal, Montreal, QC, Canada
2. Mila - Quebec AI Institute, Montreal, QC, Canada
3. Department of Neurosurgery, Faculty of Medicine and Dentistry, Palacký University Olomouc, Olomouc, Czechia
4. Department of Neurology, Faculty of Medicine and Dentistry, Palacký University Olomouc, Olomouc, Czechia
5. Center of Research in Psychology and Neuroscience (CRPN, UMR 7077), CNRS – Aix Marseille Université, Marseille, France
6. Department of Neuroimaging, Institute of Psychiatry, Psychology & Neuroscience, King's College London, London, United Kingdom
7. Wellcome Centre for Human Neuroimaging, University College London, London, United Kingdom
8. Functional Neuroimaging Unit, CRIUGM, Université de Montréal, Montreal, QC, Canada
9. Centre de Recherche du CHU Sainte-Justine, Université de Montréal, Montreal, QC, Canada

**Corresponding Author:** Jan Valošek (jan.valosek@polymtl.ca)

**ORCID:**

Jan Valošek - 0000-0002-7398-4990
Theo Mathieu - 0009-0005-7319-5288
Raphaëlle Schlienger - 0009-0004-8687-3669
Olivia Kowalczyk - 0000-0002-6543-290X
Julien Cohen-Adad - 0000-0003-3662-9532

# ABSTRACT

Precise identification of spinal nerve rootlets is relevant to delineate spinal levels for the study of functional activity in the spinal cord. The goal of this study was to develop an automatic method for the semantic segmentation of spinal nerve rootlets from T2-weighted magnetic resonance imaging (MRI) scans. Images from two open-access 3T MRI datasets were used to train a 3D multi-class convolutional neural network using an active learning approach to segment C2-C8 dorsal nerve rootlets. Each output class corresponds to a spinal level. The method was tested on 3T T2-weighted images from three datasets unseen during training to assess inter-site, inter-session, and inter-resolution variability. The test Dice score was 0.67 ± 0.16 (mean ± standard deviation across testing images and rootlets levels), suggesting a good performance. The method also demonstrated low inter-vendor and inter-site variability (coefficient of variation ≤ 1.41 %), as well as low inter-session variability (coefficient of variation ≤ 1.30 %) indicating stable predictions across different MRI vendors, sites, and sessions. The proposed methodology is open-source and readily available in the Spinal Cord Toolbox (SCT) v6.2 and higher.

# 1. INTRODUCTION

The spinal cord is a critical component of the central nervous system, containing essential motor and sensory networks and transmitting information back and forth from the brain to the peripheral nervous system. The rostrocaudal organization of the spinal cord is characterized by spinal levels (also called spinal cord segments), defined by the nerve rootlets' entry points. In contrast, the spine, which is a bony structure, is described according to vertebral levels, defined by the vertebral bodies (Diaz & Morales, 2016; Frostell et al., 2016; Kinany, Pirondini, Micera, et al., 2022). Currently, most spinal cord functional magnetic resonance imaging (MRI) studies use vertebral levels for spatial normalization to the standard space and to perform group-level analysis (Kinany, Pirondini, Micera, et al., 2022; Powers et al., 2018). To automate processing, several methods have been developed for intervertebral disc and/or vertebral level identification (Azad et al., 2021; Bozorgpour et al., 2023; Gros et al., 2018; Jamaludin et al., 2017; Mbarki et al., 2020; Rouhier et al., 2020; Ullmann et al., 2014; Vania & Lee, 2021) and spinal levels estimation using functional MRI (Kinany et al., 2020, 2024; Weber et al., 2020). However, the correspondence between vertebral and spinal levels, while described in neuroanatomy books (Standring, 2020), varies across individuals (Cadotte et al., 2015; Diaz & Morales, 2016; Mendez et al., 2021), making predictions of spinal levels based on the vertebral bodies unreliable (Cadotte et al., 2015). Previous work attempted to identify nerve rootlets based on diffusion MRI tractography (Dauleac et al., 2022; Gasparotti et al., 2013); however, these methods require the acquisition of high-resolution diffusion-weighted scans, which is not often done in clinical routine. To the best of our knowledge, no automatic tool exists for the semantic segmentation of nerve rootlets from structural (T2-weighted) MRI scans. Due to the absence of such automatic tools, researchers either have to perform manual, time-consuming and error-prone landmark identification or rely on the vertebral levels, introducing potential bias.

In this work, we introduce a method for the automatic level-specific semantic segmentation of cervical spinal cord nerve rootlets from T2-weighted MRI scans. The term *semantic segmentation* here means assigning a unique label to each class, for instance, label 2 for C2 rootlets, label 3 for C3 rootlets, and so forth. We chose turbo spin-echo T2-weighted scans for their high contrast between rootlets and cerebrospinal fluid, and for their robustness to motion compared to T1-weighted gradient-echo scans. To facilitate the creation of manual ground truth, the segmentation algorithm was trained using an active learning approach (Budd et al., 2021) on images from two open-access spinal cord MRI datasets. The proposed methodology was applied to images from multiple sources to assess the inter-rater, inter-site, inter-session, and inter-resolution variability. The segmentation method is implemented in the

`sct_deepseg` function as part of the Spinal Cord Toolbox (SCT) (De Leener et al., 2017) v6.2 and higher.

# 2. MATERIALS AND METHODS

## 2.1. Data and participants

Two open-access datasets were used for model training: *OpenNeuro ds004507* (https://openneuro.org/datasets/ds004507/versions/1.0.1, n=10) and *spine-generic multi-subject* dataset (https://github.com/spine-generic/data-multi-subject/tree/r20230223, n=267). Both datasets contain T2-weighted images acquired at 3T scanners with parameters inspired by the *Spine Generic* protocol (Cohen-Adad et al., 2021a). These images exhibit good contrast between the nerve rootlets and cerebrospinal fluid. More details about each dataset follow.

The *ds004507* dataset consists of 10 healthy participants, each scanned 3 times with different neck positions: flexion, neutral, and extension (Bédard et al., 2023). The dataset contains 0.6 mm isotropic (n=9) and 0.7 mm isotropic (n=1) T2-weighted scans. In the context of this study, we used neutral (`ses-headNormal`) and extension (`ses-headUp`) neck positions. We did not include images from the flexion neck position session (`ses-headDown`) due to poor rootlets visibility caused by neck flexion in these images. For more details on data acquisition and sequence parameters, see (Bédard et al., 2023).

The *spine-generic multi-subject* dataset consists of 267 participants scanned across 43 sites and contains 0.8 mm isotropic T2-weighted images (Cohen-Adad et al., 2021c). Participants were positioned so that the cervical spinal cord was as straight as possible to minimize partial volume effects with the surrounding cerebrospinal fluid (Cohen-Adad et al., 2021a). A subset of T2-weighted images (n=24) was used in the context of this study, as detailed in the following section. For more details on data acquisition and sequence parameters, see (Cohen-Adad et al., 2021b, 2021c).

## 2.2. Deep Learning Training Protocol

The rootlets segmentation model was trained using nnUNetv2 (Isensee et al., 2021), a popular self-configuring deep learning-based framework. All images were reoriented to the left-posterior-inferior (LPI) orientation and intensity-normalized using *z*-score normalization before training. Default data augmentation methods by nnUNetv2 were used (namely, random rotation, scaling, mirroring, Gaussian noise addition, Gaussian blurring, adjusting image brightness and contrast, simulation of low resolution, and Gamma transformation). The model was trained with a batch size of 2 using the stochastic gradient descent optimizer with a polynomial learning rate scheduler. The loss function was the sum of cross-entropy loss and Dice loss. To facilitate the creation of manual ground truth, the segmentation model was iteratively trained using an active learning approach (Budd et al., 2021), as described below. **Figure 1** shows an overview of the training pipeline.

First, we manually segmented spinal cord dorsal nerve rootlets using FSLeyes image viewer (https://fsl.fmrib.ox.ac.uk/fsl/fslwiki/FSLeyes) on 12 images (6 subjects, 2 sessions per subject) from the *ds004507* dataset. We only segmented the dorsal nerve rootlets as they are thicker relative to the ventral rootlets (Galley et al., 2021; Mendez et al., 2021) and thus easier to label. These segmentations were binary (i.e., 0: background, 1: rootlets) and used to train an initial nnUNetv2 3D model (*Model 1*, 50 epochs).

*Model 1* was used to segment the nerve rootlets on all 20 T2-weighted images from neutral and extension neck position sessions (10 subjects, 2 sessions per subject) from the *ds004507* dataset. The segmentations were visually inspected and manually corrected, if necessary. Three images were excluded due to insufficient contrast between the nerve rootlets and cerebrospinal fluid. This second set of ground truth labels (n=17) was used to train a second nnUNetv2 3D model (*Model 2*, 250 epochs).

*Model 2* was utilized to segment the nerve rootlets on T2-weighted images from the *spine-generic multi-subject* dataset. Due to the tedious nature of the manual correction process and the low contrast between the nerve rootlets and cerebrospinal fluid for some subjects, 24 T2-weighted images from this dataset were manually corrected. These 24 visually inspected and manually corrected segmentations from the *spine-generic multi-subject* dataset were combined with the previously obtained 17 segmentations from the *ds004507* dataset.

The binary ground truths were manually modified to incorporate spinal level information, assigning one label to each class (i.e., 2: C2 rootlets, 3: C3 rootlets, etc.). During this process, 5 images were excluded due to difficulties in distinguishing individual levels associated with rootlets due to their overlap.

The final dataset, consisting of 36 images with associated 7-class ground truth rootlets segmentations, was used to train the multi-class (i.e., level-specific) nnUNetv2 3D model to segment C2-C8 dorsal spinal cord nerve rootlets. We did not include the C1 dorsal rootlets as their presence varies across individuals (Diaz & Morales, 2016; Tubbs et al., 2007). The model was trained using five-fold cross-validation across 2,000 epochs per fold. An epoch is defined as one complete pass through the entire training dataset, during which the neural network processes each image in the training dataset once, adjusting its weights and biases to minimize the error in its predictions relative to the ground truth labels. Five-fold cross-validation involves training the model five times, each time with a different 80/20% training/validation split. Training/validation utilized 31 images. The testing set included 5 images (3 from *spine-generic*, 2 from *ds004507*), which were the same for all five folds to evaluate model performance across various training/validation splits. The final "production" model used all 31 images for training a "fold_all" model (100/0% training/validation split), and the same 5 images as earlier for testing. The performance of individual folds was assessed using the Dice coefficient as further described in the next section.

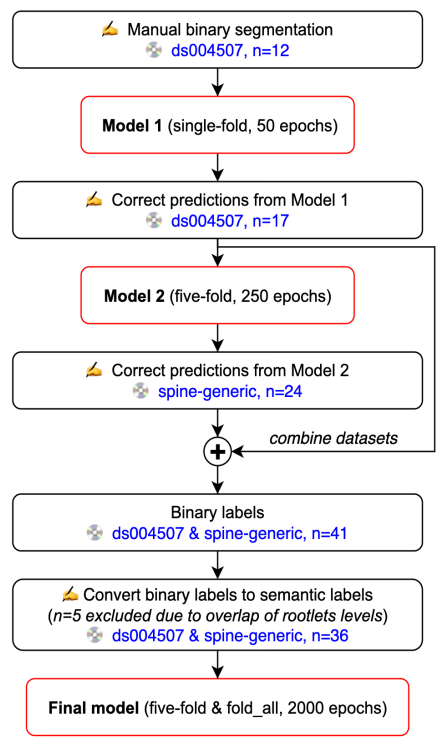

**Figure 1: Active Learning Flowchart**. The initial model (Model 1) was trained on manual binary segmentations from the *ds004507* dataset. Model 1 was used to segment additional images from the *ds004507* dataset in order to train Model 2. Then, Model 2 was utilized to segment images from the *spine-generic multi-subject* dataset. Segmentations from both datasets were combined, and binary segmentations were manually redefined into level-specific semantic segmentations (i.e., 2: C2 rootlet, 3: C3 rootlet, etc.). During the manual redefining, 5 images were excluded due to difficulties in distinguishing individual levels associated with rootlets due to their overlap resulting in 36 images used to train the Final model.

## 2.3. Inter-rater variability in nerve rootlets segmentation

To assess the inter-rater variability in manual nerve rootlets segmentation and to obtain ground truth segmentations for model testing, four raters (J.V., T.M., R.S., O.K.) from three different sites manually segmented nerve rootlets in five images (`sub-007_ses-headNormal` and `sub-010_ses-headUp` from the *ds004507* dataset and `sub-amu02`, `sub-barcelona01`, and `sub-brnoUhb03` from the *spine-generic multi-subject* dataset). Each rater manually segmented C2 to C8 dorsal nerve rootlets (i.e., with respective voxel values 2 to 8) using FSLeyes image viewer. A consensus reference segmentation mask for each image was produced using the *STAPLE* algorithm (Warfield et al., 2004). Briefly, the STAPLE algorithm iteratively computes a probabilistic estimate of the true segmentation by estimating an optimal combination of the manual segmentations from individual raters. The consensus STAPLE segmentations were considered as ground truth masks for computing the Dice coefficient at test time. The Dice coefficient was computed per level between the STAPLE ground truth and the model segmentations for each of the five test images (which were unseen during the model training/validation). Then, the mean and standard deviation (SD) Dice coefficient was calculated across spinal levels and across testing images.

**Figure 2** illustrates the procedure to obtain spinal levels from the nerve rootlets segmentation. The dorsal root entry zone was identified as the intersection of the nerve rootlets segmentation and the spinal cord segmentation (`sct_deepseg_sc` (Gros et al., 2019) dilated by 3 voxels. Then, for each rootlets level, the rostral and the caudal intersection slices were projected on the spinal cord segmentation mask to generate a semantic spinal level segmentation assigning one label to each spinal level (i.e., 2: C2 spinal level, 3: C3 spinal level, etc.).

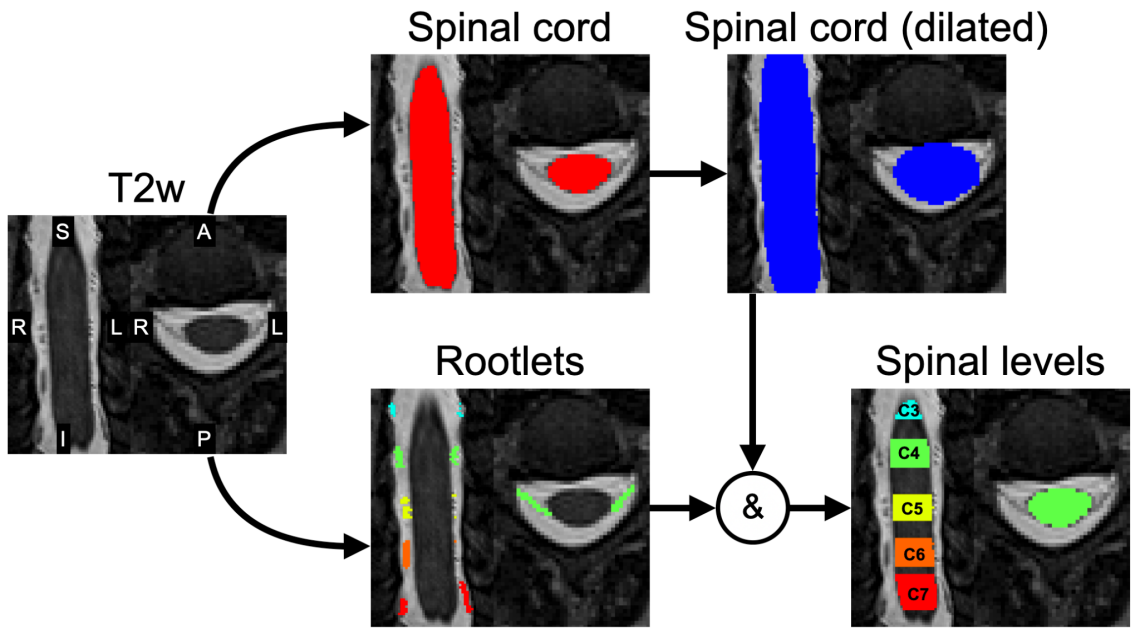

**Figure 2: Obtaining spinal levels from the nerve rootlets.** Spinal levels were identified as an intersection of dilated spinal cord segmentation and nerve rootlets segmentation.

The pontomedullary junction (PMJ) was identified using `sct_detect_pmj` (Gros et al., 2018) and visually inspected with SCT's quality report (`sct_qc`). The distance between the PMJ and the rostral, middle, and caudal slices of each spinal level was computed along the spinal cord centerline to take into account the spinal cord curvature (Bédard et al., 2023; Bédard & Cohen-Adad, 2022). The rostro-caudal length of each spinal level (i.e., the rostro-caudal rootlets distance) was calculated as a mean across five testing images.

The distance between the PMJ and the middle of each spinal level was used to calculate the inter-rater coefficient of variation (COV) for each spinal level and each testing image using the following equation:

$$COV = (Standard\ Deviation\ /\ Mean) * 100$$

The standard deviation and mean were computed across four raters for a given image and spinal level. Then, the mean COV was calculated across all testing images for each spinal level to obtain a variability for a given spinal level across testing images.

The Mann-Whitney U rank test was used to compare the spinal levels (defined as the distance between the PMJ and the middle of each spinal level) obtained from the ground truth STAPLE segmentation and spinal levels obtained using the developed segmentation model (nnUNet) for each testing image.

## 2.4. Testing Protocol

The performance of the model was tested on other 3T datasets than that used for training the model.

The first testing dataset (test-set-1) is the *spine-generic single-subject* dataset (https://github.com/spine-generic/data-single-subject), consisting of a single healthy subject (male, 38 y.o.) scanned across nineteen sites on 3T scanners from three vendors (GE, Philips, Siemens) (Cohen-Adad et al., 2021c). This dataset contains T2-weighted images at 0.8 mm isotropic (n=16) and 0.8×0.5×0.8 mm (n=3) resolution. To assess the model variability when segmenting images across different sites, we used the distance between the PMJ and the middle of each spinal level to compute the mean ± SD COV across levels and sites, separately for each vendor.

The second testing dataset (test-set-2) is the *Courtois-NeuroMod* dataset (https://github.com/courtois-neuromod/anat), an open-access dataset of healthy subjects scanned at regular intervals at the same 3T scanner (Siemens Prisma Fit) (Boudreau et al., 2023). In the context of this study, we used T2-weighted images (0.8 mm isotropic) of a single subject (`sub-01`, male, 46 y.o.) scanned ten times at regular intervals over three years. To validate the model stability in segmenting individual rootlet levels across the ten sessions, we computed the mean ± SD COV of the distance between the PMJ and the middle of each spinal level across all ten sessions for each level.

The third testing dataset (test-set-3) is a private dataset of healthy subjects (CRPN, Aix Marseille Université, Marseille, France), each with two sessions, scanned on a 3T scanner (Siemens Prisma). In the context of this study, we used T2-weighted images (0.8 mm isotropic) of four subjects (2 females, mean ± SD age: 24.5 ± 1 y.o.). To test whether the model predicted the spinal levels at the same location between sessions, we performed the pairwise comparison of the distance between the PMJ and the middle of each spinal level between sessions for each subject using the Wilcoxon signed-rank test. We also computed mean ± SD COV of the distance between the PMJ and the middle of each spinal level across sessions for each subject.

To assess model performance across different spatial resolutions, we used a single image (`sub-010_ses-headUp`) from the *ds004507* dataset (https://openneuro.org/datasets/ds004507/versions/1.0.1). The original T2-weighted isotropic 0.6 mm scan was linearly downsampled to 0.8 mm, 1.0 mm, 1.2 mm, 1.4 mm, and 1.6 mm isotropic resolutions. We computed the mean absolute error of the spinal level position between the ground truth spinal level (defined as the distance between the PMJ and the middle of each spinal level) on the 0.6 mm scan and the predicted spinal level on the downsampled scans. We also tested if there was a significant difference in the predicted levels between the 0.6 mm scan and each of the downsampled scans using the Wilcoxon signed-rank test.

## 2.5. Labeling nerve rootlets on the PAM50 template

The developed segmentation model was applied to the PAM50 T2-weighted spinal cord template image (De Leener et al., 2018) to obtain dorsal nerve rootlets in the PAM50 template space, as it could be useful for researchers. The segmented rootlets were used to estimate spinal levels, which were compared to recently updated spinal levels in the PAM50 template (Frostell et al., 2016).

# 3. RESULTS

## 3.1. Rootlets segmentation model

The level-specific nnUNetv2 3D model trained for 2,000 epochs on 31/5 training/testing T2-weighted images achieved the best performance for *fold_all* (i.e., using all 31 images for training and no image for validation). The *fold_all* achieved a test Dice score of 0.67 ± 0.16 (mean ± SD across 5 test images and all rootlets levels). The comparison of the individual folds across all rootlets levels is available in Supplementary Table 1. All further results in this paper are reported for *fold_all*. **Figure 3** shows a level-specific rootlets segmentation for a test image (`sub-barcelona01`).

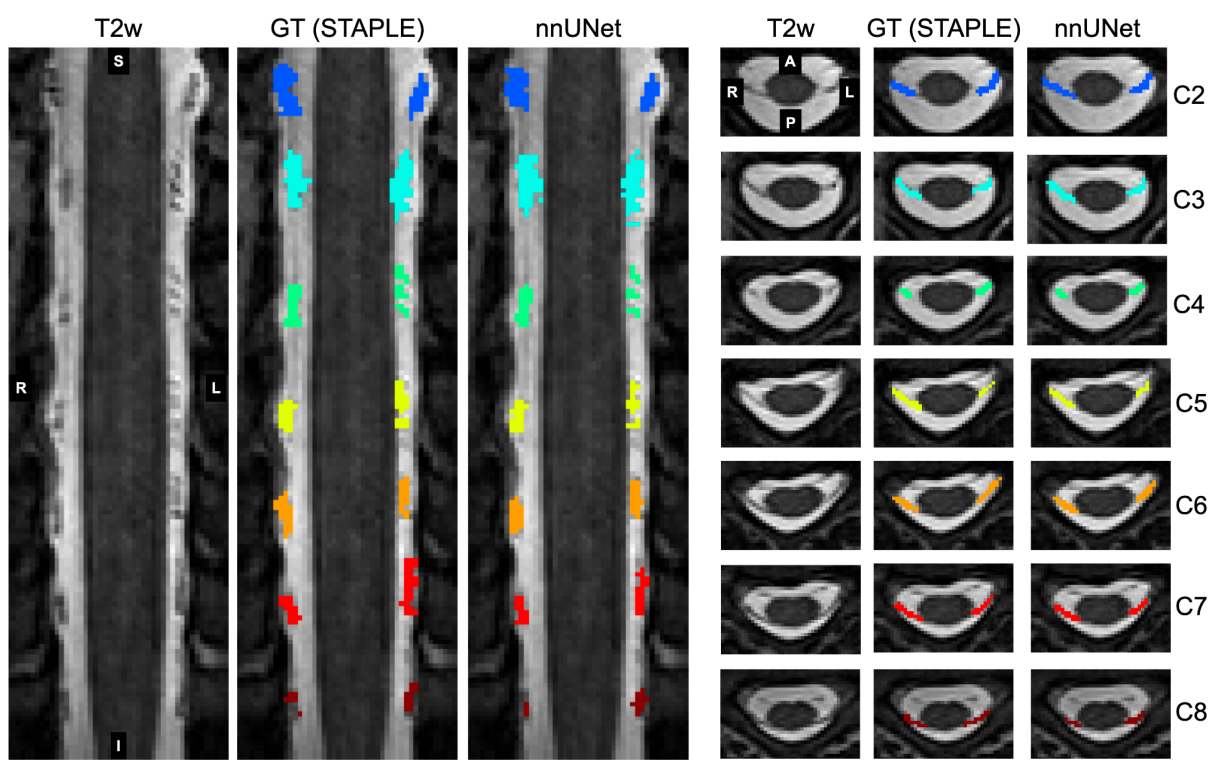

**Figure 3: Example of the semantic rootlets segmentation.** Coronal and axial views of the T2-weighted scan of a single healthy subject (`sub-barcelona01`) are overlaid with the ground truth (GT) STAPLE segmentation and the segmentation obtained using the developed segmentation model (nnUNet). Note that the coronal views were straightened using `sct_straighten_spinalcord` to show all spinal levels.

The highest test Dice score of 0.77 ± 0.08 (mean ± SD across 5 testing images) was observed for rootlets level C2, and the lowest test Dice score of 0.57 ± 0.28 was observed for rootlets level C5 (**Figure 4A**). The low Dice score for level C5 is mostly caused by an incomplete segmentation for this level for `sub-007_ses-headNormal` image due to a narrowed spinal canal (**Figure 4B**). Although some pixels were segmented for this level, they do not overlap with the ground truth STAPLE segmentation resulting in a low Dice score (**Figure 4B**). Despite the low Dice score (i.e., very little voxel overlap), the corresponding spinal level was estimated correctly (**Figure 5**). Generally, rostral rootlets (i.e., C2, C3) demonstrated higher Dice score than their caudal counterparts (i.e., C7, C8), see Discussion section for details.

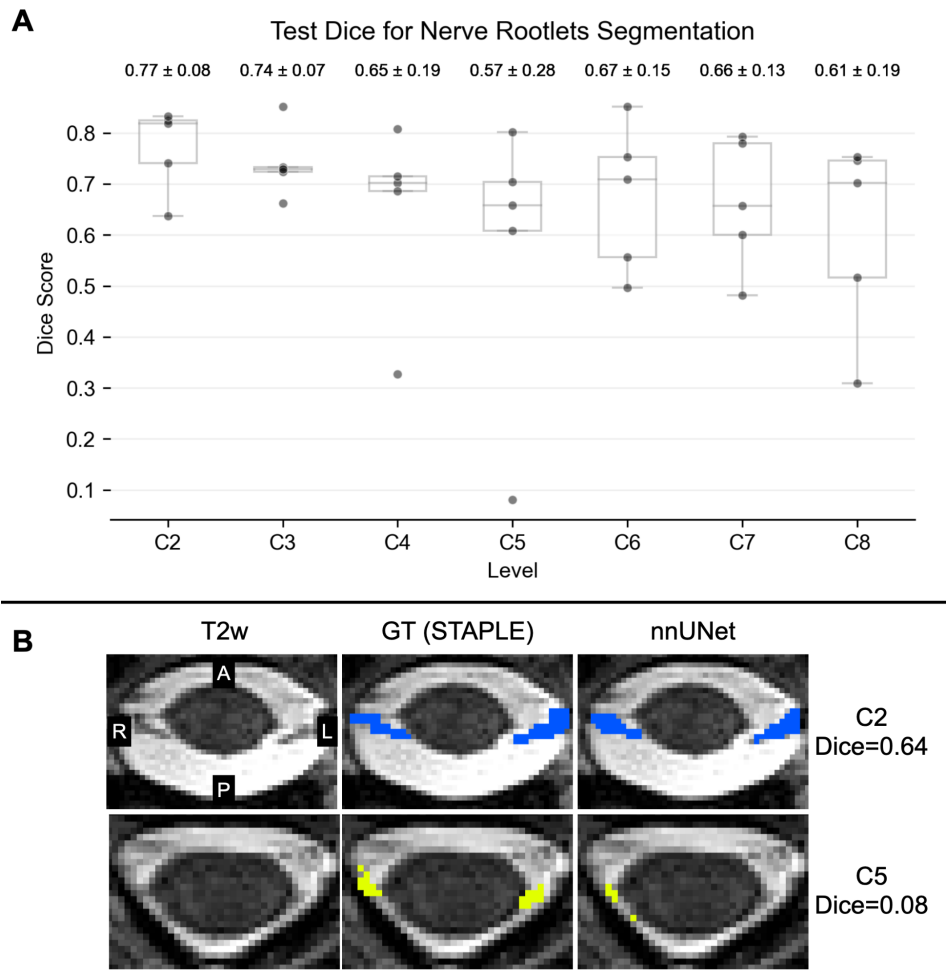

**Figure 4: Test Dice score for nerve rootlets segmentation. A)** Dice score was computed between the ground truth (GT) STAPLE segmentation and the segmentation obtained using the developed segmentation model (nnUNet). The boxplot bars represent the minima, first quartile (25%), median, third quartile (75%) and maxima. The dots represent individual testing images. The values above the boxplots represent mean ± standard deviation Dice across five testing images. **B)** Example axial slices (`sub-007_ses-headNormal`) for C2 rootlets with high Dice score (i.e., good agreement between the GT) and C5 rootlets with low Dice score (i.e., under-segmentation relative to the GT).

### 3.2. Inter-rater variability

**Figure 5** and Supplementary Table 2 show the distance from the PMJ to each spinal level for the five test images. The spinal levels were obtained as an intersection of the rootlets and dilated spinal cord segmentation (**Figure 2**) and are shown for four manual raters, the ground truth STAPLE segmentation (light gray), and the segmentation obtained using the developed segmentation model (dark gray). The distance between the PMJ and each spinal level was computed along the spinal cord centerline to take into account the spinal cord curvature. Manual raters demonstrated excellent mutual agreement (mean COV across raters and testing images ≤ 1.45 %, Supplementary Table 2) and also good correspondence with the spinal levels obtained from the nerve rootlets segmented using the developed segmentation model (**Figure 5** in dark gray). The spinal levels obtained from the ground truth STAPLE segmentation did not significantly differ from the levels obtained using the developed segmentation model (nnUNet) for any of the testing images.

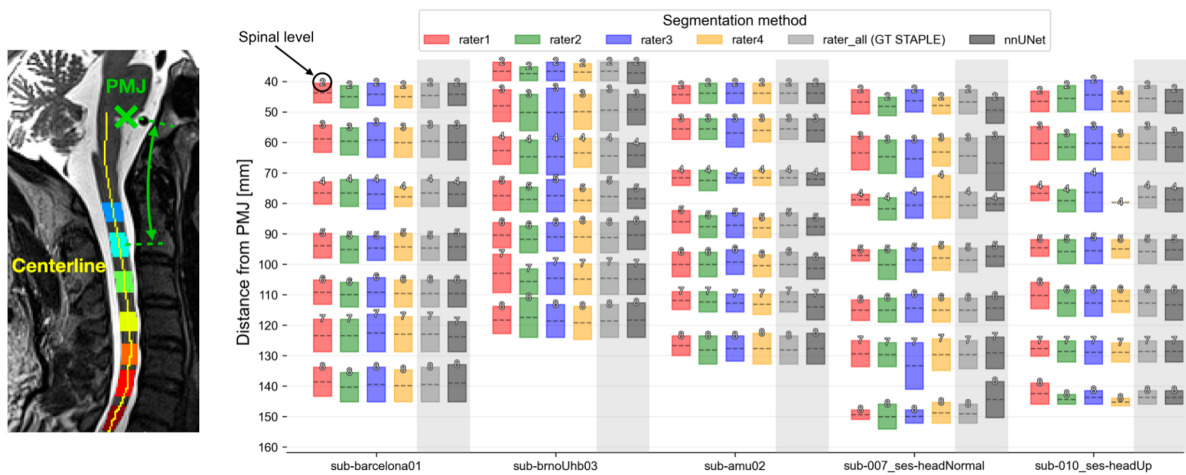

**Figure 5: Quantification of inter-rater variability and model performance for estimating spinal levels.** Nerve rootlets were manually segmented in T2-weighted images of five healthy subjects by four raters. Spinal levels obtained as an intersection of the rootlets and dilated spinal cord segmentation are shown for four manual raters, the ground truth (GT) STAPLE segmentation, and segmentation obtained using the developed segmentation model (nnUNet). Each box plot shows the mean (dashed line) and extent of the nerve rootlets segmentation intersecting with the dilated spinal cord. The identified spinal level is indicated at the top of each box.

### 3.3. Spinal level length

The rostro-caudal length of spinal levels (i.e., the rostro-caudal rootlets distance) obtained using the developed model was (mean ± SD across five testing images): C2: 7.49 ± 0.68 mm; C3: 11.30 ± 3.44 mm; C4: 6.48 ± 1.84 mm; C5: 6.89 ± 1.05 mm; C6: 8.64 ± 0.96 mm; C7: 9.10 ± 1.29 mm; and C8: 9.95 ± 2.86 mm.

## 3.4. Inter-site and Inter-session variability

This section shows the results of the model on T2-weighted images from three different datasets not used during training. For all datasets, the nerve rootlets segmentation was used to infer the spinal levels, as illustrated in **Figure 2**.

**Figure 6** shows the results of the model applied to test-set-1 (a single subject scanned across 19 sites). The obtained spinal levels correspond well across sites with an exception for C7 and C8 spinal levels for images from the `tokyoSigna1` and `tokyoSigna2` sites due to low contrast between cerebrospinal fluid and rootlets. When excluding these two sites (red crosses in **Figure 6**), the mean ± SD COV across levels and sites was 1.18 ± 0.62 % for GE, 1.35 ± 0.95 % for Philips, and 1.37 ± 0.52 % for Siemens. The mean ± SD COV across all sites (regardless of the vendors) was 1.41 ± 0.78 %.

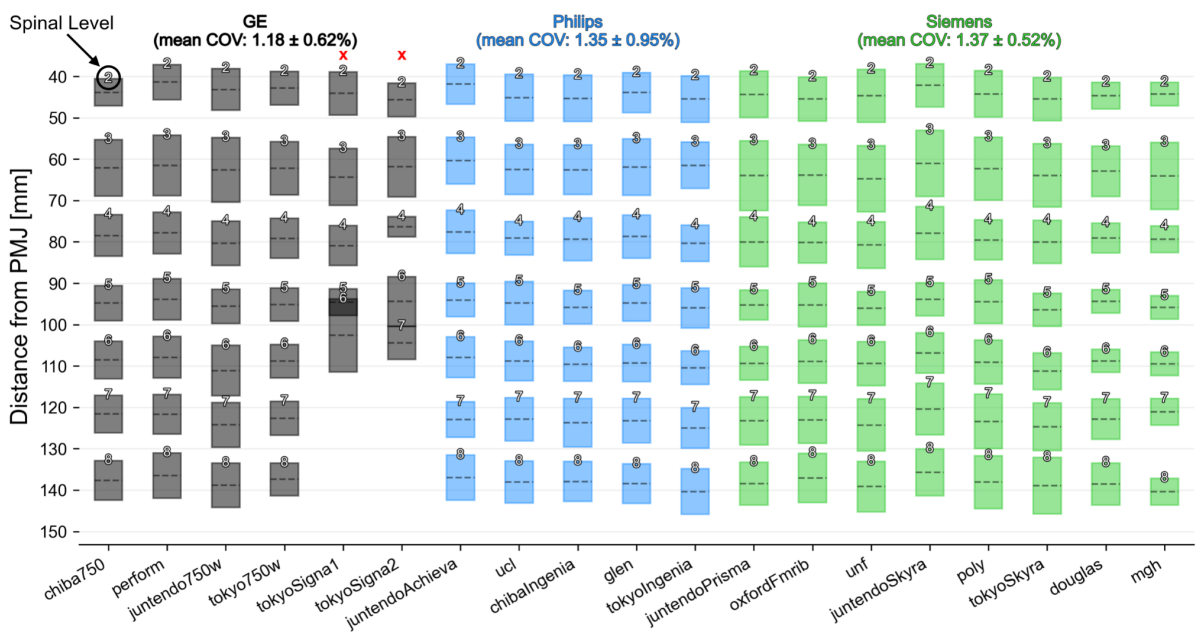

**Figure 6: Spinal level prediction on a single subject scanned across multiple sites (test-set-1).** Each box plot shows the mean (dashed line) and extent of the nerve rootlets segmentation intersecting with the dilated spinal cord. The identified spinal level is indicated at the top of each box. The plot is divided across MRI vendors (GE, Philips, Siemens), due to slight differences in acquisition parameters across vendors. The `tokyoSigna1` and `tokyoSigna2` sites were excluded from statistics (red crosses) due to low contrast between cerebrospinal fluid and rootlets at C7 and C8 spinal levels.

**Figure 7** shows the results of the model applied to test-set-2 (a single subject scanned ten times over three years on the same scanner). The obtained spinal levels show good agreement across all ten sessions with a COV of 1.30 ± 0.50 % (mean ± SD across ten sessions and all levels). The highest inter-session COV of 2.29 % was observed for rootlets level C2, and the lowest inter-session COV of 0.77 % was observed for rootlets level C6.

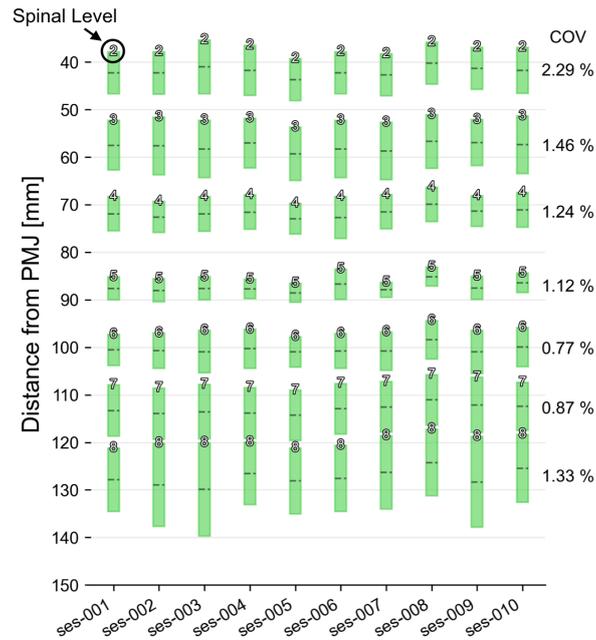

**Figure 7: Spinal level prediction on a single subject scanned over three years at the same site (test-set-2).** Each box plot shows the mean (dashed line) and extent of the nerve rootlets segmentation intersecting with the dilated spinal cord. The identified spinal level is indicated at the top of each box. Inter-session coefficient of variation (COV) for each spinal level is shown on the right.

**Figure 8** shows the results of the model applied to test-set-3 (four healthy subjects, each scanned two times on the same 3T scanner). The spinal levels exhibited excellent correspondence between sessions (i.e., there was no statistically significant difference in the distance between the PMJ and the middle of each spinal level between sessions for any subject) with inter-session COV < 2.56 %.

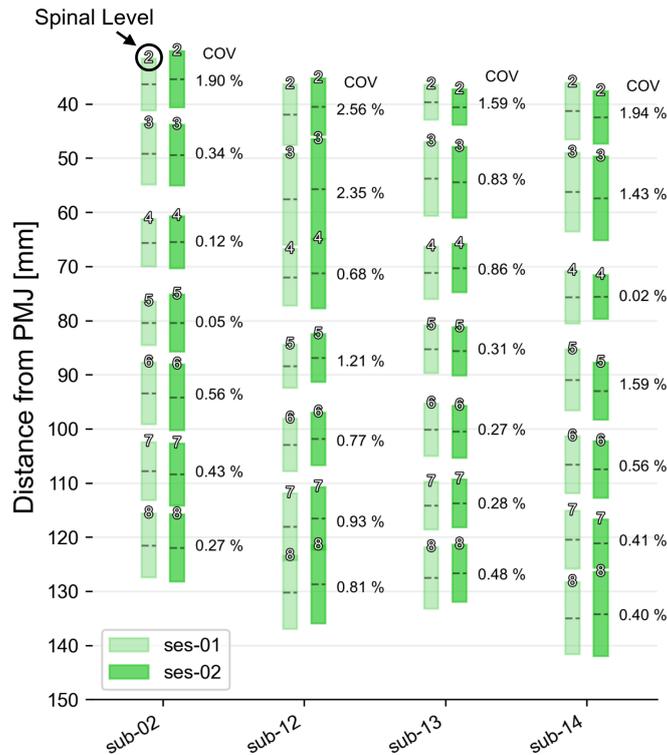

**Figure 8: Spinal level prediction on four subjects with scan/rescan experiments at the same site (test-set-3).** Each box plot shows the mean (dashed line) and extent of the nerve rootlets segmentation intersecting with the dilated spinal cord. The identified spinal level is indicated at the top of each box. Inter-session coefficient of variation (COV) for each spinal level is shown on the right for each subject.

### 3.5. Inter-resolution variability

**Figure 9** illustrates the testing of the developed model on a T2-weighted image at various isotropic resolutions: 0.6 mm (original), 0.8 mm, 1.0 mm, 1.2 mm, 1.4 mm, and 1.6 mm. The mean absolute error across spinal levels for each resolution using 0.6 mm as a reference was: 5.60 mm, 3.54 mm, 3.77 mm, 4.44 mm and 0.58 mm for the downsampled images at 0.8 mm, 1.0 mm, 1.2 mm, 1.4 mm, and 1.6 mm, respectively. The Wilcoxon signed-rank test revealed statistically significant differences ($p < .05$) in predicting spinal levels between the 0.6 mm scan and the 0.8 mm, 1.0 mm, 1.2 mm, and 1.4 mm downsampled scans. The predicted spinal levels between 0.6 mm and 1.6 mm scans did not differ significantly.

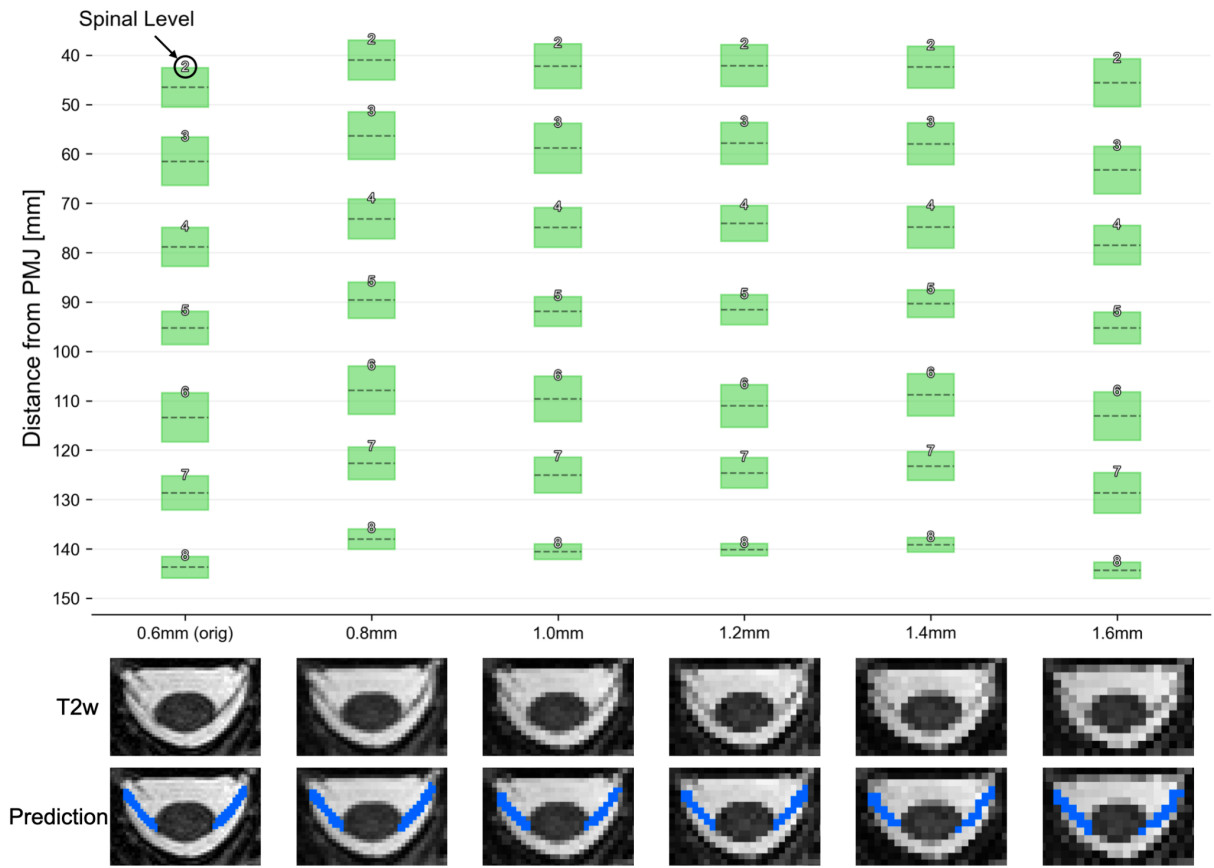

**Figure 9: Model performance across spatial resolutions.** Each box plot at the top panel shows the mean (dashed line) and extent of the nerve rootlets segmentation intersecting with the dilated spinal cord. The identified spinal level is indicated at the top of each box. The bottom panel shows an example slice from the C3 level across resolutions.

## 3.6. Labeling nerve rootlets on the PAM50 template

**Figure 10** illustrates nerve rootlets and spinal levels obtained by our method on the PAM50 T2-weighted template image together with the spinal levels based on (Frostell et al., 2016). The percentage overlap between the spinal levels estimated by our method and the spinal levels based on (Frostell et al., 2016) was: C2: 37.79%; C3: 67.67%; C4: 58.50%; C5: 52.63%; C6: 54.49%; C7: 50.02%; and C8: 75.49%.

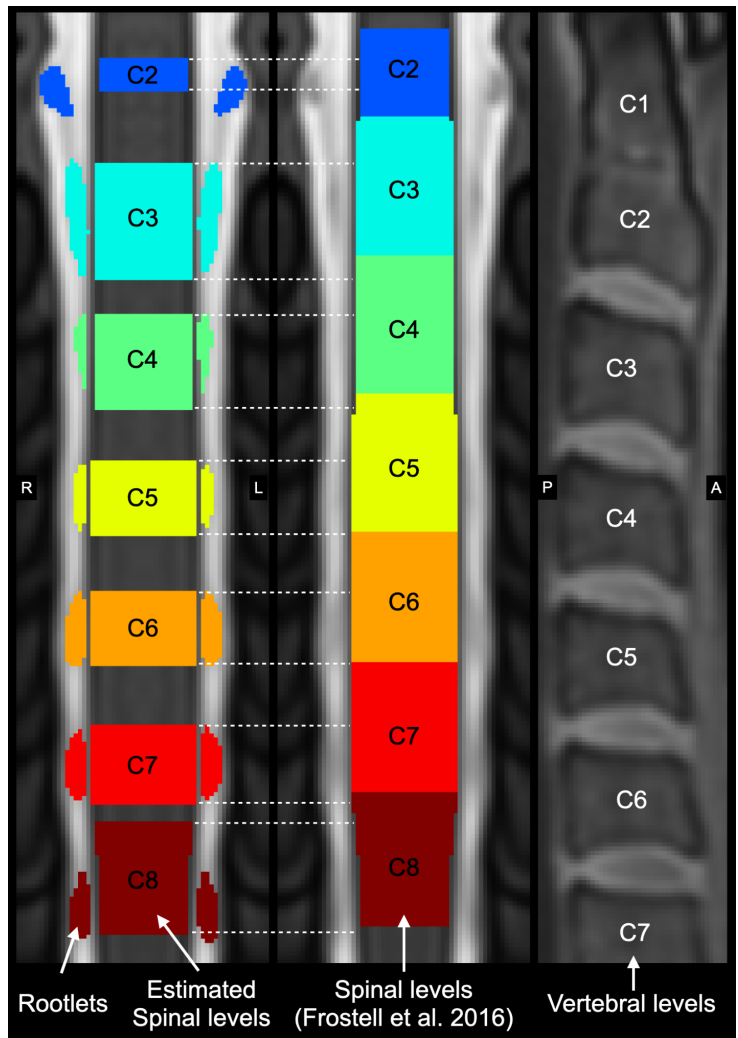

**Figure 10: Nerve rootlets and spinal levels on the PAM50 template space.** The developed model was applied to the PAM50 T2-weighted template image to segment dorsal cervical rootlets and estimate spinal levels (left panel). The middle panel shows spinal levels based on (Frostell et al., 2016) and their correspondence with the estimated levels (dashed white lines). The right panel shows vertebral levels of the PAM50 T1-weighted image.

# 4. DISCUSSION

This study introduced a deep learning-based model for the automatic segmentation of spinal cord nerve rootlets from T2-weighted MRI scans and a method for identifying the corresponding spinal levels based on the rootlets segmentation. The proposed methodology showed low variability across different MRI vendors, sites, and sessions indicating robust segmentation performance.

## 4.1. Nerve rootlets segmentation

The segmentation of spinal cord nerve rootlets is challenging due to their submillimeter size and varying anatomy across levels. In any single slice (axial, coronal, or sagittal) of a volumetric MRI image, the rootlets only appear as a few connected voxels emerging from the spinal cord and going through the spinal canal to exit via the intervertebral foramen grouped as spinal nerves. Therefore, a 3D acquisition protocol, ideally with isotropic resolution, is desirable to capture the rootlets' spatial arrangement and minimize the partial volume effect (Diaz & Morales, 2016).

As the spinal cord is shorter than the spine, and the C2 rootlets project quasi-orthogonally onto the spinal cord, the more caudal the rootlets are located, the more oblique their angle relative to the spinal cord to reach the corresponding foramina (Mendez et al., 2021). When the angle is important (typically starting from the thoracic levels downwards), rootlets from more than one level can overlap, making the disentanglement of each nerve rootlet extremely difficult, even at very high resolution. This is the reason why, in the present study, we only considered the cervical levels and did not include images with overlapping rootlets levels for training the segmentation model.

Despite these challenges, the segmentation model achieved a good performance, with a test Dice score of 0.67 ± 0.16 (mean ± SD across 5 test images and all rootlets levels computed against the STAPLE ground truths). A consequence of the angulation of the rootlets mentioned earlier is that the model performed better on rostral rootlets (C2, C3) than the more caudal ones (C7, C8). Other causes of lower performance in the more caudal levels include a higher spinal cord curvature, reduced spinal canal space, and a higher prevalence of signal dropout possibly caused by flow effects (Cohen-Adad et al., 2021a).

Because of the time-consuming nature of manual rootlets annotation, our final training/validation set only comprised 31 images, which is relatively small compared to the number of images used to train other medical image segmentation models, typically over 100

(Gros et al., 2019; Lemay et al., 2021). There are a few reasons for the reasonably good performance (test Dice of 0.67) despite a relatively small sample size. First, the training was done on good-quality images with well-visible rootlets for all levels. Second, compared to tasks such as segmentation of microbleeds (Qi Dou et al., 2016), intramedullary tumors (Lemay et al., 2021), or spinal cord injury lesions (Naga Karthik et al., 2024), where the size, location, and number of the regions to segment vary, here, the consistency of the task (i.e., always the same number of classes, geometrically stable within the image and between classes) makes the tasks somewhat easier to learn. Third, the nnU-Net pipeline included aggressive data augmentation strategies, which effectively inflated the number of input patches used for model training, which in some segmentation tasks (such as here, where the geometry of labels is fairly consistent across subjects) proved to be a very effective means to obtain a performant model at low data regime (Shorten & Khoshgoftaar, 2019).

The model also showed good generalizability on three external test datasets (unseen during the model training), featuring notably low inter-vendor and inter-site variability (**Figure 6**), as well as low inter-session variability (**Figure 7** and **Figure 8**). It is important to note, however, that the demographics (young, healthy volunteers) and the image parameters (isotropic turbo spin-echo T2-weighted sequence) of the external test datasets were similar to the datasets used for the model training. As further discussed later, applying the model to out-of-distribution data is a potential avenue for future research.

## 4.2. Identification of spinal levels from the nerve rootlets segmentation

The level-specific spinal nerve rootlets segmentation can be used for the identification of the spinal levels, which serve as an alternative coordinate system to the commonly used vertebral levels defined based on the vertebral bodies (Diaz & Morales, 2016; Frostell et al., 2016; Kinany, Pirondini, Micera, et al., 2022). Spinal levels are particularly relevant for functional MRI studies, which would benefit from spinal levels for spatial normalization to the spinal cord standard space for group-level analysis (Kinany et al., 2024; Kinany, Pirondini, Mattera, et al., 2022; Kinany, Pirondini, Micera, et al., 2022; Seifert et al., 2023; Weber et al., 2016, 2020). In this study, we proposed a method for spinal level identification based on the intersection of rootlets segmentation and dilated spinal cord segmentation. To reliably identify intersections between these two segmentations, we dilated the spinal cord mask by 3 voxels. While this method provides an approximate spinal level estimation, it relies on the quality of the spinal cord mask and would benefit from further validation and improvements (see section *Limitations and future directions*). Previous studies, involving manual measurements of spinal level lengths from high-resolution MRI scans (0.4×0.4×0.3 mm) or cadavers, reported that cervical levels have similar lengths (Cadotte et al., 2015; Mendez et al., 2021). Cadotte's MRI study measured cervical rootlets group lengths from 9.4 ± 1.4 mm to 10.5 ± 2.2 mm (Cadotte et al., 2015), while Mendez's cadaveric study obtained lengths ranging from 9.4 ± 1.9 mm to 12.4 ± 2.0 mm (Mendez et al., 2021). In our study, the level lengths varied from 6.48 ± 1.84 mm to 11.30 ± 3.44 mm, which is a slightly larger range than what was previously reported. This variation could be attributed to differences in image resolutions, and, more likely, the accuracy of our proposed automatic method to estimate spinal levels, as further discussed in section _Limitations and future directions_.

### 4.3. Labeling nerve rootlets on the PAM50 template

Nerve rootlets visible in the PAM50 template were fairly well segmented (despite a slight under-segmentation), and the estimated spinal levels align reasonably well—although not perfectly—with those from (Frostell et al., 2016) (**Figure 10**). The slight misalignment could be caused by the imperfect segmentation of the nerve rootlets, which itself could be caused by the relatively high angulation of rootlets below the C4 spinal level, making it more difficult to distinguish them from the spinal cord. Another likely cause for the imperfect segmentation is that the PAM50 template was created by averaging images from 50 adult subjects with different morphometry (De Leener et al., 2018). Even though the template creation pipeline involved co-registering all these images, it is possible that the nerve rootlets were not perfectly aligned across all subjects, causing some blurriness in the nerve rootlets and hence sub-optimal segmentation performance.

### 4.4. MRI contrast to visualize nerve rootlets

We chose heavily weighted turbo spin-echo T2-weighted scans for their high contrast between rootlets and cerebrospinal fluid, and for their robustness to motion compared to T1-weighted gradient-echo sequences (Branco et al., 2023; Cohen-Adad et al., 2021a). Given that rootlets are subtle structures covering only a few voxels, the robustness to the motion artifacts is a crucial consideration. Additionally, in the available open-access spinal cord databases that follow the _Spine Generic_ protocol (Cohen-Adad et al., 2021a), T2-weighted images have higher spatial resolution than T1-weighted scans (0.8 mm versus 1 mm). Other sequences exist that showed good results in visualizing nerve rootlets, for example, dual-echo steady-state (Galley et al., 2021) or multi-echo gradient-echo scans (Cohen-Adad et al., 2022).

### 4.5. Limitations and future directions

The present model only segments the cervical spinal nerve rootlets, mostly for the reasons mentioned earlier (difficulty isolating rootlets in the thoraco-lumbar spinal cord due to their higher angulation, and less spinal canal space). The higher resolution and signal-to-noise ratio offered by ultra-high field systems (7T and above) open the door to identifying nerve rootlets across a larger rostro-caudal coverage (Zhao et al., 2014).

The model only segments the dorsal rootlets. While there is a good spatial agreement between the ventral and the dorsal rootlets (in terms of rostro-caudal location), differences in their anatomy exist (Mendez et al., 2021), and in some experimental setups (e.g., paradigms looking at motor vs. sensory responses), researchers might want to be able to isolate ventral from dorsal rootlets. The decision to focus only on the dorsal rootlets was practical: they are thicker relative to the ventral rootlets (Galley et al., 2021; Mendez et al., 2021) and thus easier to label. Also, we did not include the C1 dorsal rootlets due to the variability in their presence across individuals (Diaz & Morales, 2016; Tubbs et al., 2007). Future studies can enrich the present model by adding these missing labels.

The method introduced to estimate the spinal level can also be further improved. Taking the intersection between the nerve rootlets segmentation and the dilated spinal cord mask is problematic in that while it works reasonably well for the rostral rootlets that project quasi-orthogonally to the cord (e.g. C2), for more angulated caudal rootlets, the intersection with the dilated mask likely shifts caudally the estimated spinal levels with respect to its *true* location. The decision for a 3 mm dilation was also purely arbitrary and was based on preliminary results comparing various kernel sizes, producing either no intersection (kernel too small) or too much caudal shift of the estimated spinal level (kernel too large). Moreover, the proposed method to estimate spinal level implies that there could be a variable gap between adjacent spinal levels, depending on how the rootlets extend along the superior-inferior direction. There is no consensus about the presence of a gap: while some studies documented spinal levels without gaps between levels (Frostell et al., 2016), other cadaveric (Mendez et al., 2021) and in vivo MRI (Cadotte et al., 2015) do report a gap. Future studies could focus on estimating the spinal level by taking advantage of the existing PAM50 template and its now available nerve rootlets segmentation. For example, we could imagine a method where the nerve rootlets of a subject would be registered to that of the PAM50 in order to warp PAM50 spinal levels back to the subject native space. That registration can take multiple forms, for example, a straightening of the cord, followed by a rootlet-wise alignment using a translation along the rostro-caudal axis, potentially aided by a regularization penalty term across all rootlet-based registration.

Results showed that the input image resolution might have an impact on the location of the predicted spinal levels (see **Figure 9**). However, the mean absolute error of the spinal level position measured as a distance from the PMJ remained within a reasonable range across the downsampled scans (3.54 mm to 5.60 mm). These results need to be taken with a grain of salt, however, given the difficulty in properly segmenting the nerve rootlets. For example, it is possible that the spinal levels for the so-called 'reference' resolution (i.e., 0.6 mm) that was used to compute the mean absolute error and the statistical test were in fact more 'wrong' than the predictions based on downsampled resolutions.

Future research could focus on increasing the training dataset size (including more contrasts and resolutions), assessing the model performance across different neck positions since rootlets' angulation changes across neck flexion/extension, testing the model on 7T images, and applying it to pathologies and the pediatric population.

## 5. CONCLUSION

This study introduced an automatic method for the semantic segmentation of spinal cord nerve rootlets on MRI scans and a method to estimate spinal level based on the nerve rootlets segmentation. The developed method demonstrated low variability across different MRI vendors, sites, and sessions indicating robust segmentation performance. The segmentation model is open-source, available in SCT v6.2 and higher and can be used to inform functional MRI studies and other studies where spinal level information is relevant.

# ETHICS

All datasets used in this study complied with all relevant ethical regulations.

# DATA AND CODE AVAILABILITY

The data used for the segmentation model development come from open-access datasets and can accessed at https://openneuro.org/datasets/ds004507/versions/1.0.1 and https://github.com/spine-generic/data-multi-subject/tree/r20230223.

The processing scripts used in this study are available at: https://github.com/ivadomed/model-spinal-rootlets/tree/r20240129. The packaged and ready-to-use segmentation model can be applied to custom data via the `sct_deepseg` function using the newly added flag `-task seg_spinal_rootlets_t2w` as part of the Spinal Cord Toolbox (SCT) v6.2 and higher: https://github.com/spinalcordtoolbox/spinalcordtoolbox/tree/6.2. The nerve rootlets segmentation in the PAM50 space is available in SCT v6.2 and higher.

# AUTHOR CONTRIBUTIONS

J.V.: Data Curation, Formal Analysis, Funding acquisition, Investigation, Methodology, Visulasion, and Writing (original draft, review & editing). T.M.: Data Curation, Formal Analysis, Investigation, Methodology, Visulasion, and Writing (original draft, review & editing). R.S., O.K.: Data Curation, Investigation, Methodology, and Writing (review & editing). J.C.A.: Conceptualization; Data Curation, Formal Analysis, Funding acquisition, Investigation, Methodology, Supervision, Visulasion, and Writing (review & editing).

# DECLARATION OF COMPETING INTERESTS

The authors declared no potential conflicts of interest with respect to the research, authorship, and/or publication of this article.

# ACKNOWLEDGEMENTS

We thank Naga Karthik Enamundram and Sandrine Bédard for valuable feedback, fruitful discussions, and review of the manuscript. We also thank Nick Guenther and Mathieu Guay-Paquet for their assistance with the management of the datasets, and Joshua Newton and Mathieu Guay-Paquet for their contributions in helping us implement the algorithm to SCT. We also thank the reviewers for their insightful comments.

# FUNDING

Funded by the Canada Research Chair in Quantitative Magnetic Resonance Imaging [CRC-2020-00179], the Canadian Institute of Health Research [PJT-190258], the Canada Foundation for Innovation [32454, 34824], the Fonds de Recherche du Québec - Santé [322736, 324636], the Natural Sciences and Engineering Research Council of Canada [RGPIN-2019-07244], the Canada First Research Excellence Fund (IVADO and TransMedTech), the Courtois NeuroMod project, the Quebec BioImaging Network [5886, 35450], INSPIRED (Spinal Research, UK; Wings for Life, Austria; Craig H. Neilsen Foundation, USA), Mila - Tech Transfer Funding Program. This project has received funding from the European Union's Horizon Europe research and innovation programme under the Marie Skłodowska-Curie grant agreement No 101107932 and is supported by the Ministry of Health of the Czech Republic, grant nr. NU22-04-00024.